\documentclass[10pt,twocolumn,letterpaper]{article}

\usepackage[letterpaper,margin=0.75in,columnsep=0.28in]{geometry}
\usepackage[T1]{fontenc}
\usepackage{amsmath,amssymb}
\usepackage{mathptmx}  
\usepackage{graphicx}
\usepackage{float}
\usepackage{cuted}
\usepackage{booktabs}
\usepackage{threeparttable} 
\usepackage{array}
\usepackage{caption}
\usepackage{xcolor}
\usepackage{tikz}
\usetikzlibrary{positioning, arrows, arrows.meta, shapes.geometric, calc}
\usepackage[hidelinks]{hyperref}
\usepackage{flafter}   
\usepackage{cite}      
\usepackage{titlesec}
\usepackage{dblfloatfix}  
\titleformat{\section}{\fontsize{12}{14}\selectfont\bfseries}{\thesection.}{0.5em}{}
\titleformat{\subsection}{\fontsize{11}{13}\selectfont\bfseries}{\thesubsection.}{0.5em}{}
\titleformat{\subsubsection}{\fontsize{10}{12}\selectfont\bfseries}{\thesubsubsection.}{0.5em}{}
\titlespacing*{\section}{0pt}{2.2ex plus 0.6ex minus .2ex}{1.2ex plus .2ex}
\titlespacing*{\subsection}{0pt}{2.0ex plus 0.5ex minus .2ex}{1.0ex plus .2ex}
\titlespacing*{\subsubsection}{0pt}{1.8ex plus 0.5ex minus .2ex}{0.9ex plus .2ex}

\renewenvironment{abstract}
  {\par\vspace{0.5em}{\centering\fontsize{12}{14}\selectfont\bfseries\abstractname\par}\vspace{0.5em}}
  {\par}

\newcommand{\Vprev}{V_{\mathrm{prev}}}
\newcommand{\Sint}{S_{\mathrm{int}}}
\newcommand{\zint}{z_{\mathrm{int}}}
\newcommand{\vint}{v_{\mathrm{int}}}
\newcommand{\CRGB}{C_{\mathrm{RGB}}}
\newcommand{\CDepth}{C_{\mathrm{Depth}}}
\newcommand{\pmax}{p_{\max}}
\newcommand{\Tramp}{T_{\mathrm{ramp}}}

\title{\textbf{MSNN-LINet:}\\ Cross-Modal Learning via Continuous Linear Integration}
\author{
  Gabriel Clinger\\
  The George Washington University\\
  \texttt{g.clinger@gwmail.gwu.edu}
}
\date{}

\begin{document}

\maketitle

\begin{abstract}
We present \textbf{LINet} (Linear Integration Network), a Multi-Stream
Neural Network (MSNN) for RGB-D scene classification. Current multi-modal
architectures treat feature fusion as a discrete, ad-hoc event: early
fusion entangles representations prematurely, late fusion isolates them
until the final layer, and hybrid or attention-based methods require
architectural guesswork to place intermediate fusion blocks. LINet
addresses this structural compromise by maintaining three dedicated
parallel streams (RGB, depth, and integration) where a novel
\textbf{Linear Integration Convolution (LIConv2d)} operator enables
continuous cross-modal learning at every layer. The integration stream
receives raw filtered signals from both modality streams and combines them
before the nonlinear activation threshold, conceptually inspired by somatic
integration preceding the neuronal firing decision.

Implementing continuous integration exposes a critical initialization
pathology: Kaiming initialization of the bridging weights scrambles
gradients before they reach the stream backbones, producing a failure mode
that resembles overfitting but is corrupted gradient flow. A $1/N$ constant
initialization mitigates this. We employ \textbf{progressive modality
dropout}, a curriculum adapted to continuous fusion in which blanking
probability increases from zero, preventing pathway collapse, a form of
negative co-learning, by forcing robust independent stream representations.
Trained from scratch on SUN RGB-D 19-class scene classification, LINet
reaches 45.2\% mean class accuracy at ResNet18 scale, outperforming prior
from-scratch results, and rises to 49.6\% with in-domain RGB-D (ScanNet)
pretraining.
\end{abstract}

\section{Introduction}
Multi-modal classification requires models to balance domain-specific feature
extraction with cross-modal representation learning. However, the dominant
architectural paradigms (including early, late, and intermediate hybrid fusion)
predominantly rely on discrete, ad-hoc fusion events. This forces a fundamental
structural compromise regarding where and when modalities should interact.
\textbf{Early fusion} concatenates depth as a fourth input channel, processing
the fused tensor through a single backbone: parameter-efficient but entangling.
A single filter set must simultaneously learn color texture, lighting
invariance, and geometric structure, with no mechanism to develop
modality-specific expertise \cite{wangA2016, songX2017}.
\textbf{Late fusion} trains separate RGB and depth backbones, combining
predictions at the output: clean modality separation, but at the cost of doubled
parameters, roughly doubled training time, and no cross-modal learning during
feature extraction. Neither stream can inform the other while representations are
built \cite{wangW2018}.

\begin{figure}[H]
\centering
\resizebox{\columnwidth}{!}{%
\begin{tikzpicture}[>=stealth, font=\sffamily]
\tikzstyle{std_node}=[circle, draw=black, very thick, minimum size=1.5cm, fill=gray!10]
\tikzstyle{macro_rgb}=[rectangle, draw=red!80, fill=red!10, very thick, minimum width=2.2cm, minimum height=0.9cm, align=center, rounded corners=2pt]
\tikzstyle{macro_int}=[rectangle, draw=green!60!black, fill=green!10, very thick, minimum width=2.2cm, minimum height=0.9cm, align=center, rounded corners=2pt]
\tikzstyle{macro_dep}=[rectangle, draw=blue!80, fill=blue!10, very thick, minimum width=2.2cm, minimum height=0.9cm, align=center, rounded corners=2pt]
\tikzstyle{soma}=[circle, draw=black, very thick, inner sep=0pt, minimum size=0.9cm, fill=white]
\tikzstyle{act}=[rectangle, draw=black, very thick, fill=white, rounded corners=2pt, minimum height=0.9cm, minimum width=1.5cm, align=center]
\node[anchor=west, font=\Large\bfseries] at (-1, 11) {A. Traditional Neuron Connection};
\node[std_node] (A_N1) at (0, 9) {$N_{l-1}$};
\node[std_node] (A_N2) at (5, 9) {$N_l$};
\draw[->, line width=2pt] (A_N1) -- node[above, font=\bfseries] {Single Weight ($W$)} (A_N2);
\node[anchor=west, font=\Large\bfseries] at (9.5, 11) {B. MSNN Multi-Weight Connection};
\node[std_node] (B_N1) at (10.5, 9) {$N_{l-1}$};
\node[std_node] (B_N2) at (16.5, 9) {$N_l$};
\draw[->, line width=2pt, draw=red!80] (B_N1.45) to[out=30, in=150] node[above, font=\bfseries] {RGB Weight ($W_{\mathrm{rgb}}$)} (B_N2.135);
\draw[->, line width=2pt, draw=green!60!black] (B_N1.0) -- node[fill=white, font=\bfseries, inner sep=2pt] {Self-Weight ($\Vprev$)} (B_N2.180);
\draw[->, line width=2pt, draw=blue!80] (B_N1.-45) to[out=-30, in=-150] node[below, font=\bfseries] {Depth Weight ($W_{\mathrm{depth}}$)} (B_N2.-135);
\node[anchor=west, font=\Large\bfseries] at (-1, 6.5) {C. Inside the MSNN Neuron: Somatic Integration};
\node[rectangle, draw, very thick, rounded corners=12pt, minimum width=10.5cm, minimum height=5.5cm, fill=gray!5, dashed] (C_Box) at (7.5, 2) {};
\node[anchor=south, font=\Large\bfseries] at (C_Box.north) {MSNN Neuron ($N_l$)};
\draw[very thick, dashed, draw=gray] (B_N2.210) -- (C_Box.north west);
\draw[very thick, dashed, draw=gray] (B_N2.330) -- (C_Box.north east);
\node[std_node] (C_N1) at (0.5, 2) {$N_{l-1}$};
\node[macro_rgb] (C_Wrgb) at (4, 3.8) {Dendrite 1\\$W_{\mathrm{rgb}}$};
\node[macro_int] (C_Wint) at (4, 2) {Self-Weight\\$\Vprev$};
\node[macro_dep] (C_Wdep) at (4, 0.2) {Dendrite 2\\$W_{\mathrm{depth}}$};
\draw[->, line width=1.5pt, draw=red!80] (C_N1.45) to[out=30, in=180] (C_Wrgb.west);
\draw[->, line width=1.5pt, draw=green!60!black] (C_N1.0) -- (C_Wint.west);
\draw[->, line width=1.5pt, draw=blue!80] (C_N1.-45) to[out=-30, in=180] (C_Wdep.west);
\node[soma] (C_Soma) at (8, 2) {\Large $\bigoplus$};
\node[act] (C_Act_rgb) at (10.5, 3.8) {Activation};
\node[act] (C_Act_int) at (10.5, 2) {Activation};
\node[act] (C_Act_dep) at (10.5, 0.2) {Activation};
\draw[->, line width=1.5pt, draw=red!80, dashed] (C_Wrgb.east) to[out=0, in=110] node[pos=0.4, left, font=\bfseries] {$V_1$} (C_Soma.110);
\draw[->, line width=1.5pt, draw=blue!80, dashed] (C_Wdep.east) to[out=0, in=250] node[pos=0.4, left, font=\bfseries] {$V_2$} (C_Soma.250);
\draw[->, line width=1.5pt, draw=green!60!black] (C_Wint.east) -- (C_Soma.west);
\draw[->, line width=1.5pt] (C_Soma.east) -- node[above, font=\bfseries] {$\zint$} (C_Act_int.west);
\draw[->, line width=2pt, draw=red!80] (C_Wrgb.east) -- (C_Act_rgb.west);
\draw[->, line width=2pt, draw=red!80] (C_Act_rgb.east) -- (14.5, 3.8) node[right, font=\bfseries, text=black] {To $N_{l+1}$};
\draw[->, line width=2pt, draw=green!60!black] (C_Act_int.east) -- node[above, font=\bfseries, text=black] {$h_l$} (14.5, 2) node[right, font=\bfseries, text=black] {To $N_{l+1}$};
\draw[->, line width=2pt, draw=blue!80] (C_Wdep.east) -- (C_Act_dep.west);
\draw[->, line width=2pt, draw=blue!80] (C_Act_dep.east) -- (14.5, 0.2) node[right, font=\bfseries, text=black] {To $N_{l+1}$};
\end{tikzpicture}%
}
\caption{The MSNN multi-weight architecture vs.\ a traditional network.
\textbf{(A)} A standard layer-to-layer connection uses a single fused weight.
\textbf{(B)} MSNN replaces it with parallel modality-specific weights spanning
the same transition. \textbf{(C)} Inside the MSNN neuron, modality filters
($W_{\mathrm{rgb}}, W_{\mathrm{depth}}$) feed independent activations while also branching through
learned projections ($V_1, V_2$) into a pre-activation integration
($\bigoplus$).}
\label{fig:msnn_vs_traditional}
\end{figure}
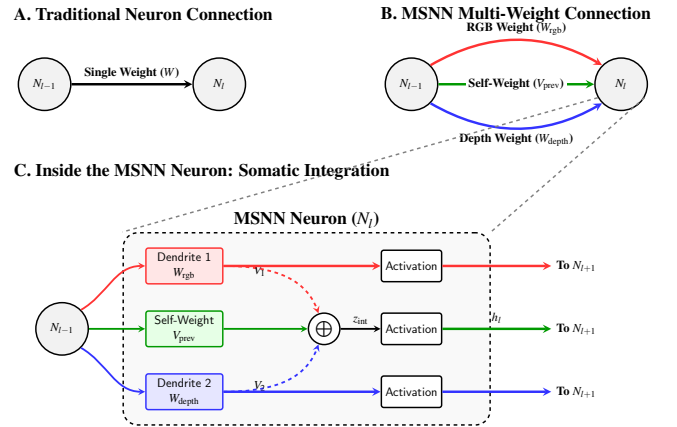

The primate visual system navigates this same trade-off through a principled
architecture: the parvocellular and magnocellular pathways process color/texture
and depth/motion signals through dedicated, structurally separate channels, yet
continuously fuse these representations through extensive cross-talk and
integration within the visual cortex \cite{nassi2009}. Critically, this
integration operates at the soma on \emph{pre-threshold} dendritic signals: each
pathway filters its modality through dedicated dendritic arbors, and the soma
combines these raw filtered signals before deciding whether to fire. This is the
architectural principle we operationalize through \textbf{Multi-Stream Neural
Networks (MSNNs)} (Figure~\ref{fig:msnn_vs_traditional}). We introduce
\textbf{LINet}, an MSNN built on ResNet18 with a novel LIConv2d operator that
maintains three concurrent streams, $S_1$ (RGB), $S_2$ (depth), and $\Sint$
(integration), with continuous cross-modal integration at every layer.

Implementing continuous, layer-by-layer integration introduces training
pathologies not addressed by standard initialization practices; resolving them
requires both architectural and optimization solutions. Our main contributions
are:

\begin{itemize}
  \item We propose the LIConv2d operator enabling continuous pre-activation
  integration across parallel network streams (Section~\ref{sec:liconv}).
  \item We identify and mitigate the integration gradient corruption problem:
  standard Kaiming initialization of bridging $1\times1$ convolutions scrambles
  gradients before they reach the stream backbones. A deterministic $1/N$
  constant initialization restores gradient coherence, enabling the stream
  backbones to learn and the network to generalize
  (Section~\ref{sec:init}, ablation in Section~\ref{sec:ablation}).
  \item We design a progressive modality dropout schedule tailored to continuous
  fusion and demonstrate, via ablative stream monitoring, that it reverses
  negative co-learning, transitioning the modality streams from brittle
  co-dependence to robust independent feature extractors.
\end{itemize}

\section{Related Work}
\label{sec:related}

\subsection*{Multi-Modal Feature Fusion:}
Feed-forward multi-modal classification relies on discrete, ad-hoc events such as
early, late, or intermediate fusion. Early fusion concatenates raw inputs,
entangling modality representations before they can specialize \cite{wangA2016,
songX2017}. Late fusion isolates modalities in separate
backbones, preventing cross-modal learning during feature extraction; notably,
Wang, W. and Neumann \cite{wangW2018} demonstrated that fixed CNN geometry strictly prevents
early depth information from informing RGB feature propagation. Intermediate
precursors such as RFBNet \cite{deng2019} and MMSS \cite{wangA2016}
recognized the need for multi-level sharing but implemented it via post-hoc
operations on independently pre-trained encoders, meaning bottom-layer extractors
never see cross-modal gradients. None of these methods maintain dedicated
parallel modality pathways while enabling continuous cross-modal learning at
every layer.

LINet's integration stream addresses this by participating at every convolutional
layer from the first training step, receiving pre-activation signals before any
nonlinearity. This continuous integration exposes unique initialization and
training dynamics (Sections~\ref{sec:init}--\ref{sec:progmd}) absent from
block-level designs. While recent Transformer-based architectures achieve high
accuracy through complex cross-modal attention \cite{ahmed2025}, their
intricate routing obscures foundational fusion mechanics. We validate LINet on a
standard ResNet backbone to isolate and analyze the
gradient dynamics and feature contributions of the LIConv2d operator, free from
the confounding variables of attention biases. However, the mathematical
formulation is backbone-agnostic and extends naturally to future attention-based
architectures or mixture-of-experts gating.

\subsection*{Isolating Fusion from Stream Initialization:}
A parallel line of work answers a different question. Faced with a small RGB-D
corpus and no ImageNet-scale labeled depth dataset, these methods ask how best to
obtain a usable depth stream, whether by training the depth network from scratch
on the target set \cite{songX2017, ayub2020}, translating RGB features into the
depth stream \cite{du2019}, or self-supervising a shared encoder on the RGB-D
pairs themselves \cite{yang2023}, or pretraining the backbone directly on RGB-D
pairs rather than RGB alone (DFormer~\cite{yin2024}), the latter two both noting
that RGB-pretrained depth weights stay biased toward RGB and yield a suboptimal
solution. Their contributions live in this stream-initialization machinery; the
fusion step itself is typically a late combination of independently formed
features, so what is varied across these works is how each stream is seeded, not
how the streams interact. LINet instead targets the interaction itself. By seeding both modalities
identically, the only variable that distinguishes our model from this baseline is
that fusion happens continuously and is itself learned at every layer. Holding
stream initialization fixed isolates the contribution of the integration
mechanism, rather than confounding it with an asymmetric improvement to the depth
stream.

\subsection*{Modality Dropout and Negative Co-Learning:}
Modality dropout, stochastically masking entire input streams during training,
was introduced as a regularizer against modality dominance \cite{neverova2015}
and subsequently combined with contrastive objectives \cite{gu2025}.
Recent extensions diverge along two axes: robustness-oriented frameworks
\cite{nezakati2024, zhang2025} use learned placeholder tokens to
hallucinate absent modalities at inference, while curriculum and adaptive
approaches \cite{chlon2025, magal2025} dynamically vary masking
schedules to reverse negative co-learning (NCL), where joint training suppresses
unimodal capacity. The current frontier moves beyond manual scheduling entirely:
entropy-gated frameworks \cite{chlon2025} drive the dropout mask from
training-time entropy, allowing the network to determine on a per-batch basis
when a modality should be dropped. These adaptive methods address intra-dataset
variance that fixed schedules cannot (not all images are equally difficult) but
require auxiliary classifiers to evaluate semantic-level confidence at the fusion
point.

Our progressive schedule employs modality dropout to force independent stream
representations, improving training dynamics. LINet's continuous per-layer fusion collapses the standard
distinction between input-level and embedding-level masking: zeroing the raw
stream input algebraically zeroes the contribution at every subsequent
integration point. The progressive ramp from zero is architecturally motivated,
allowing integration weights to form before independence is demanded; we find
that LINet's continuous integration is nonetheless sufficiently robust that
static dropout from epoch zero remains viable, a property we attribute to the
architectural mechanisms discussed in Section~\ref{sec:streamhealth}. Notably,
AECF \cite{chlon2025} independently adopts an identical linear ramp for
entropy-gate regularization; in LINet, this schedule shape is instead driven by
integration weight dynamics (Section~\ref{sec:ablation}).

\begin{figure*}[t]
\centering
\includegraphics[width=\textwidth]{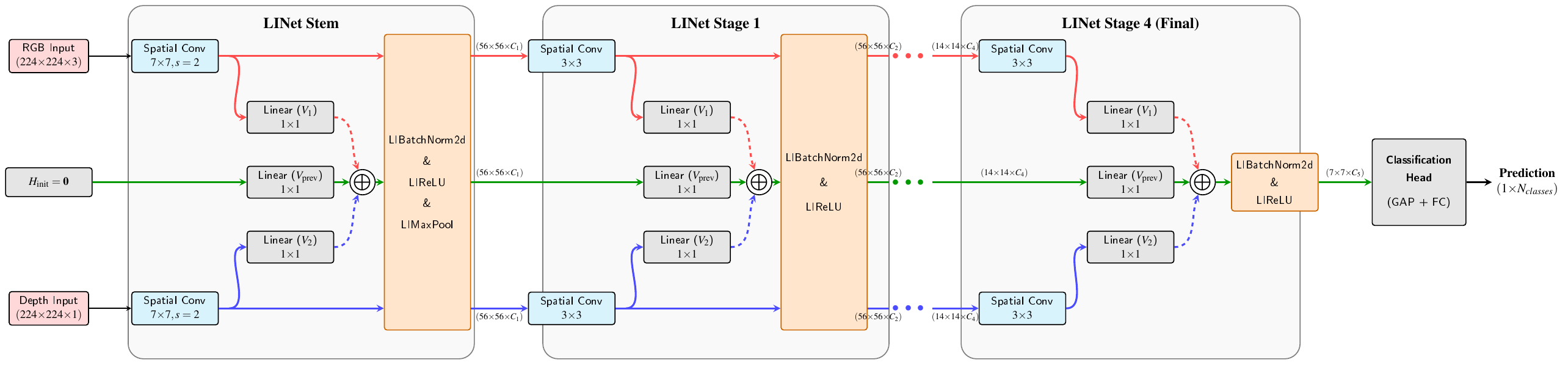}
\caption{LINet: model architecture.}
\label{fig:linet_end_to_end}
\end{figure*}

\section{Methodology: The LINet Architecture}

In a standard single-modality CNN like ResNet, a semantic manifold is iteratively
constructed by projecting important features from the previous layer into a
deeper, increasingly abstract dimensional space. LINet extends this foundational
principle to multiple modalities through a dedicated central integration stream.
At each convolutional block, this central stream acts as a continuous multimodal
ledger. It takes the accumulated manifold from the previous layer, smoothly
projected via $1\times1$ convolutions to maintain structural cohesion, and
directly fuses it with the newly extracted, modality-pure features from the
parallel RGB and Depth pathways.

By continuously executing this $\mathrm{Integrated}_{\mathrm{prev}} +
\mathrm{RGB}_{\mathrm{current}} + \mathrm{Depth}_{\mathrm{current}}$ integration, the architecture
avoids the pitfalls of both early and late fusion. Instead, it builds an evolving,
shared semantic space that, while mirroring the layer-by-layer conceptual
progression of a standard ResNet, projects each modality into it through its own
learned mapping.

LINet is built on the ResNet18 skeleton (Figure~\ref{fig:linet_end_to_end}), with
all standard components replaced by their Linear Integration counterparts:
Conv2d~$\rightarrow$~LIConv2d, BatchNorm2d~$\rightarrow$~LIBatchNorm2d,
BasicBlock~$\rightarrow$~LIBasicBlock, etc. Each LI module maintains three
parallel streams, \textbf{$S_1$ (RGB)}, \textbf{$S_2$ (Depth)}, and
\textbf{$\Sint$ (Integration)}, following ResNet18's channel progression
$[64, 64, 128, 256, 512]$ (scaled by width factor $\alpha = 0.75$ in our
experiments to $[48, 48, 96, 192, 384]$; see Section~\ref{sec:impl}) with no
shared weights at any layer. At the stem, $\Sint$ is computed from the raw stream
outputs ($V_1{*}d_1 + V_2{*}d_2$) with the $\Vprev$ term omitted by design, as
there is no prior integration state at the first layer. The remainder of the
architecture follows the standard ResNet18 structure; the linear classifier
attaches exclusively to $\Sint$, which carries the fused multi-modal
representation.

To formalize the forward propagation let
$\mathcal{X}_{\mathrm{rgb}} \in \mathbb{R}^{3 \times H \times W}$ and
$\mathcal{X}_{\mathrm{depth}} \in \mathbb{R}^{1 \times H \times W}$ represent the paired
input modalities. The network initialization begins with modality-specific stem
convolutions to generate the initial feature spaces, while the integration stream
is initialized as a zero-tensor prior to the first layer's summation:
\begin{equation*}
x_1^{(0)} = \mathcal{S}_{\mathrm{rgb}}(\mathcal{X}_{\mathrm{rgb}}), \quad
x_2^{(0)} = \mathcal{S}_{\mathrm{depth}}(\mathcal{X}_{\mathrm{depth}}), \quad
h^{(0)} = \mathbf{0}
\end{equation*}

For a network of depth $L$ (representing the sequential blocks of our shared
ResNet18 backbone), the hidden states at any layer $l \in \{1, \dots, L\}$ are
computed recursively by our multi-stream function $\mathcal{F}_{\mathrm{LI}}$ (defined by
the routing in Section~\ref{sec:liconv}):
\begin{equation}
(x_1^{(l)}, h^{(l)}, x_2^{(l)}) =
\mathcal{F}_{\mathrm{LI}}^{(l)}\!\left(x_1^{(l-1)}, h^{(l-1)}, x_2^{(l-1)}\right)
\end{equation}

Upon exiting the final convolutional block $L$, the spatial dimensions of all
three streams are collapsed via Global Average Pooling (GAP) to produce distinct
1D feature vectors:
\begin{equation*}
\resizebox{0.92\columnwidth}{!}{$\displaystyle
v_{\mathrm{rgb}} = \mathrm{GAP}(x_1^{(L)}), \quad
\vint   = \mathrm{GAP}(h^{(L)}), \quad
v_{\mathrm{depth}} = \mathrm{GAP}(x_2^{(L)})
$}
\end{equation*}

To compute the final classification probabilities, the integration stream's
feature vector $\vint$ is projected into the class space (e.g.,
$\mathbb{R}^{19}$ for standard SUN-RGBD scene categories) via a fully connected
classification weight matrix $W_{\mathrm{cls}}$:
\begin{equation}
\begin{aligned}
y &= W_{\mathrm{cls}} \cdot \vint + b_{\mathrm{cls}}\\
\hat{p} &= \mathrm{Softmax}(y)
\end{aligned}
\end{equation}

This formulation ensures that the final predictive distribution $\hat{p}$ is
conditioned exclusively on the integration stream's continuously fused
cross-modal representation ($\vint$), which has aggregated information from both
modality-specific pathways at every layer.

\subsection{Linear Integration Convolution (LIConv2d)}
\label{sec:liconv}
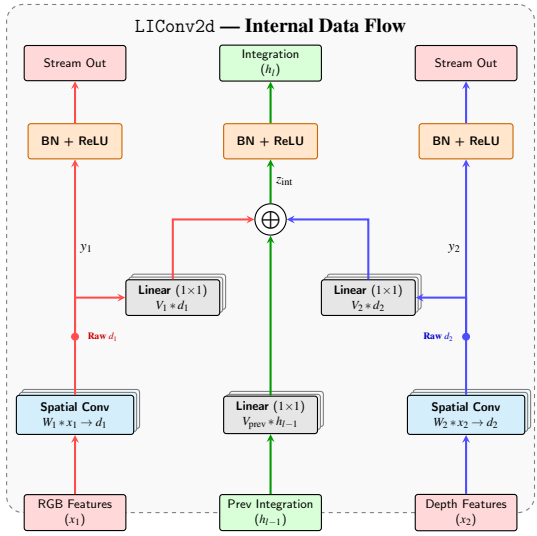
\begin{figure}[H]
\centering
\resizebox{0.82\columnwidth}{!}{%
\begin{tikzpicture}[
    >=stealth, font=\sffamily\small,
    pink_box/.style={rectangle, draw=black, thick, fill=red!15, minimum width=2.6cm, minimum height=0.9cm, align=center, rounded corners=2pt},
    blue_box/.style={rectangle, draw=black, thick, fill=cyan!15, minimum width=3.0cm, minimum height=1.0cm, align=center, rounded corners=2pt},
    gray_box/.style={rectangle, draw=black, thick, fill=gray!20, minimum width=2.4cm, minimum height=0.9cm, align=center, rounded corners=2pt},
    green_box/.style={rectangle, draw=black, thick, fill=green!15, minimum width=2.6cm, minimum height=0.9cm, align=center, rounded corners=2pt},
    yellow_box/.style={rectangle, draw=orange!80!black, thick, fill=orange!20, minimum width=2.4cm, minimum height=0.9cm, align=center, rounded corners=2pt},
    sum_node/.style={circle, draw=black, thick, fill=white, inner sep=0pt, minimum size=0.8cm},
    bias_node/.style={text=black, font=\small\bfseries, align=center},
    math_anno/.style={font=\normalsize\bfseries, text=black, align=center},
    rgb_arrow/.style={->, draw=red!70, line width=1.5pt},
    rgb_line/.style={-, draw=red!70, line width=1.5pt},
    dep_arrow/.style={->, draw=blue!70, line width=1.5pt},
    dep_line/.style={-, draw=blue!70, line width=1.5pt},
    int_arrow/.style={->, draw=green!60!black, line width=1.5pt},
    int_line/.style={-, draw=green!60!black, line width=1.5pt}
]
\node[rectangle, draw=gray, thick, rounded corners=12pt, fill=gray!5, minimum width=13.5cm, minimum height=13cm, dashed] (layer_bg) at (5, 6.5) {};
\node[anchor=north, font=\Large\bfseries, inner sep=10pt] at (layer_bg.north) {\texttt{LIConv2d} --- Internal Data Flow};
\node[pink_box]  (in_rgb) at (0, 0)  {RGB Features\\($x_1$)};
\node[green_box] (in_int) at (5, 0)  {Prev Integration\\($h_{l-1}$)};
\node[pink_box]  (in_dep) at (10, 0) {Depth Features\\($x_2$)};
\node[blue_box, draw=gray, fill=cyan!5]  at (0.15, 2.65) {};
\node[blue_box, draw=gray, fill=cyan!10] at (0.075, 2.575) {};
\node[blue_box] (conv_rgb) at (0, 2.5) {\textbf{Spatial Conv}\\$W_1 * x_1 \rightarrow d_1$};
\node[blue_box, draw=gray, fill=cyan!5]  at (10.15, 2.65) {};
\node[blue_box, draw=gray, fill=cyan!10] at (10.075, 2.575) {};
\node[blue_box] (conv_dep) at (10, 2.5) {\textbf{Spatial Conv}\\$W_2 * x_2 \rightarrow d_2$};
\node[gray_box, draw=gray, fill=gray!5]  at (5.15, 2.65) {};
\node[gray_box, draw=gray, fill=gray!10] at (5.075, 2.575) {};
\node[gray_box] (v_prev) at (5, 2.5) {\textbf{Linear} ($1{\times}1$)\\$\Vprev * h_{l-1}$};
\draw[rgb_arrow] (in_rgb.north) -- (conv_rgb.south);
\draw[dep_arrow] (in_dep.north) -- (conv_dep.south);
\draw[int_arrow] (in_int.north) -- (v_prev.south);
\coordinate (fork_rgb) at (0, 4.5);
\coordinate (fork_dep) at (10, 4.5);
\draw[rgb_line] (conv_rgb.north) -- (fork_rgb);
\draw[dep_line] (conv_dep.north) -- (fork_dep);
\filldraw[red!70]  (fork_rgb) circle (2.5pt);
\filldraw[blue!70] (fork_dep) circle (2.5pt);
\node[text=red!80!black,  font=\scriptsize\bfseries, align=left,  anchor=west] at (0.2, 4.5)  {Raw $d_1$};
\node[text=blue!80!black, font=\scriptsize\bfseries, align=right, anchor=east] at (9.8, 4.5)  {Raw $d_2$};
\node[gray_box, draw=gray, fill=gray!5]  at (2.65, 5.65) {};
\node[gray_box, draw=gray, fill=gray!10] at (2.575, 5.575) {};
\node[gray_box] (v1) at (2.5, 5.5) {\textbf{Linear} ($1{\times}1$)\\$V_1 * d_1$};
\node[gray_box, draw=gray, fill=gray!5]  at (7.65, 5.65) {};
\node[gray_box, draw=gray, fill=gray!10] at (7.575, 5.575) {};
\node[gray_box] (v2) at (7.5, 5.5) {\textbf{Linear} ($1{\times}1$)\\$V_2 * d_2$};
\draw[rgb_arrow] (fork_rgb) |- (v1.west);
\draw[dep_arrow] (fork_dep) |- (v2.east);
\node[sum_node]  (sum_soma) at (5, 7.5) {\Large $\bigoplus$};
\draw[rgb_arrow] (v1.north) |- (sum_soma.180);
\draw[dep_arrow] (v2.north) |- (sum_soma.0);
\draw[int_arrow] (v_prev.north) -- (sum_soma.south);
\node[yellow_box] (bn_rgb)  at (0, 9.5)  {\textbf{BN + ReLU}};
\node[yellow_box] (bn_soma) at (5, 9.5)  {\textbf{BN + ReLU}};
\node[yellow_box] (bn_dep)  at (10, 9.5) {\textbf{BN + ReLU}};
\draw[rgb_arrow] (fork_rgb) -- node[right, math_anno] {$y_1$} (bn_rgb.south);
\draw[dep_arrow] (fork_dep) -- node[left,  math_anno] {$y_2$} (bn_dep.south);
\draw[int_arrow] (sum_soma.north) -- node[right, math_anno] {$\zint$} (bn_soma.south);
\node[pink_box]  (out_rgb) at (0, 11.5)  {Stream Out};
\node[green_box] (out_int) at (5, 11.5)  {Integration\\($h_l$)};
\node[pink_box]  (out_dep) at (10, 11.5) {Stream Out};
\draw[rgb_arrow] (bn_rgb.north)  -- (out_rgb.south);
\draw[int_arrow] (bn_soma.north) -- (out_int.south);
\draw[dep_arrow] (bn_dep.north)  -- (out_dep.south);
\end{tikzpicture}%
}
\caption{Internal data flow of the \texttt{LIConv2d} module. Each modality is
filtered by its spatial filter ($W_1, W_2$); each raw output $d_i$ branches both
to a stream activation and, via $V_i$, to the somatic summation, after which
independent BatchNorm+ReLU pathways yield the parallel outputs ($y_1$, $h_l$,
$y_2$). All convolutions use \texttt{bias=False}; BatchNorm's affine parameters
act as the per-channel offset.}
\label{fig:liconv2d_vertical}
\end{figure}

Conceptually inspired by the dendritic integration principle, the LIConv2d
operator (Figure~\ref{fig:liconv2d_vertical}) consists of three sequential steps:

\begin{itemize}
  \item Modality-specific filtering: each modality stream processes inputs
  independently via a learned spatial filter $W_i$ ($3\times3$ for stream
  convolutions).
  \item Pre-activation integration: raw filtered signals $d_i$ are combined via
  learned $1\times1$ mixing weights $V_i$ before the activation function.
  \item Activation: ReLU fires exclusively on the already-integrated signal,
  implementing integrate-then-activate.
\end{itemize}

Formally, for a layer with $C_{\mathrm{in}}$ input and $C_{\mathrm{out}}$ output channels, the
LIConv2d weight tensors are:

\begin{itemize}
  \item $W_1 \in \mathbb{R}^{C_{\mathrm{out}} \times C_{\mathrm{in}} \times 3\times3}$: spatial
  filter for $S_1$ (RGB stream)
  \item $W_2 \in \mathbb{R}^{C_{\mathrm{out}} \times C_{\mathrm{in}} \times 3\times3}$: spatial
  filter for $S_2$ (Depth stream)
  \item $V_1 \in \mathbb{R}^{C_{\mathrm{out}} \times C_{\mathrm{out}} \times 1\times1}$: mixing
  weight for $S_1 \rightarrow$ integration
  \item $V_2 \in \mathbb{R}^{C_{\mathrm{out}} \times C_{\mathrm{out}} \times 1\times1}$: mixing
  weight for $S_2 \rightarrow$ integration
  \item $\Vprev \in \mathbb{R}^{C_{\mathrm{out}} \times C_{\mathrm{in}} \times 1\times1}$: state
  projection weight
\end{itemize}

The forward computation per LIConv2d layer:
\begin{equation}
\resizebox{0.9\columnwidth}{!}{$\displaystyle
\begin{aligned}
d_1     &= W_1 *_s x_1 && \text{\scriptsize modality filtering, } S_1 \text{ (stride } s) \\
d_2     &= W_2 *_s x_2 && \text{\scriptsize modality filtering, } S_2 \text{ (stride } s) \\
\zint   &= V_1{*}d_1 + V_2{*}d_2 + \Vprev{*}_s h_{l-1} && \text{\scriptsize pre-activation integration} \\
h_l     &= \mathrm{ReLU}(\mathrm{BN}(\zint)) && \text{\scriptsize integration output} \\
y_1     &= \mathrm{ReLU}(\mathrm{BN}(d_1)) && \text{\scriptsize } S_1 \text{ output} \\
y_2     &= \mathrm{ReLU}(\mathrm{BN}(d_2)) && \text{\scriptsize } S_2 \text{ output}
\end{aligned}
$}
\end{equation}

At transition layers (stride $s=2$), $\Vprev$ applies the same stride as the main
convolution, simultaneously halving spatial resolution and mixing channels,
analogous to the $1\times1$ projection shortcut in standard ResNet downsampling
blocks. Unlike the square modality mixing weights ($V_1, V_2$) $\Vprev$ must
project the previous layer's $C_{\mathrm{in}}$ channels to the current layer's $C_{\mathrm{out}}$
channels during spatial downsampling. $V_1$ and $V_2$ always use stride$=1$ since
$d_1$ and $d_2$ are already at the target resolution. Integration operates on raw
modality outputs before stream normalization and activation, so the integration
pathway sees pure filtered signals. All spatial filters and integration
convolutions omit independent bias terms, as the affine parameters of the
subsequent BatchNorm layers serve as the effective per-channel offset throughout.

\subsection{The Integration Initialization Problem}
\label{sec:init}
Kaiming initialization, the standard convention for ResNet architectures,
scrambles the gradients of the stream-to-integration bridging weights
($V_1, V_2$) before they reach the modality-specific backbones. This gradient
corruption prevents backbone representations from forming, creating a failure
mode that mimics overfitting. We mitigate this using a dense constant fill,
initializing every entry of the $V_1$ and $V_2$ $[C_{\mathrm{out}}, C_{\mathrm{out}}, 1, 1]$
tensors to $1/N$ ($= 0.5$ for $N=2$):

\begin{center}
\noindent\resizebox{\columnwidth}{!}{\ttfamily init.constant\_(integration\_from\_streams[i], 1.0 / num\_streams)}
\end{center}

\begin{figure}[H]
\centering
\includegraphics[width=\columnwidth]{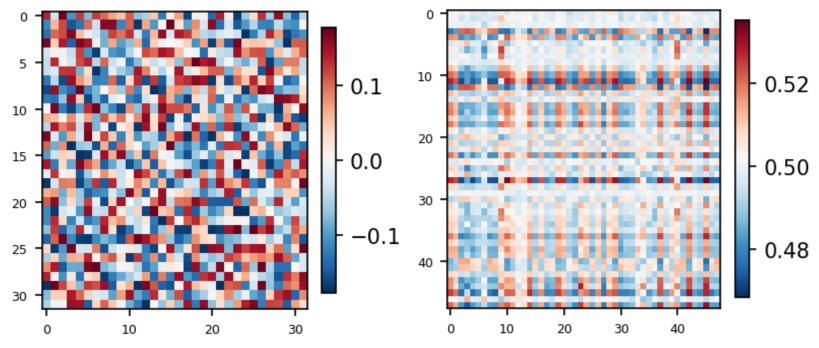}
\caption{$V_1$ integration weight matrix. Left: Kaiming initialization, weights
fail to develop channel-specific structure. Right: $1/N$ initialization, weights
develop channel-specific routing patterns.}
\label{fig:init_weights}
\end{figure}

Although this causes significant forward-pass amplification (up to $192\times$), BN immediately absorbs it. BN is scale-invariant. The decisive
property is \textbf{gradient coherence}: every output channel receives the same
initial linear combination, so the stream backbones receive identical, stable
signals from the first update. $V_1$ and $V_2$ then break symmetry naturally via
channel-specific gradients from the Kaiming-initialized $W_1$ and $W_2$ (see
Figure~\ref{fig:init_weights}).

$\Vprev$ receives different treatment: orthogonal initialization for square layers
($C_{\mathrm{in}}=C_{\mathrm{out}}$), which preserves gradient norms \cite{saxe2014}, and
Kaiming$\times0.1$ for non-square transition layers. This reflects $\Vprev$'s
distinct role: carrying the integration stream's own history forward rather than
mixing external streams.

\subsection{Progressive Modality Dropout}
\label{sec:progmd}
To prevent the network from over-relying on a single modality, we apply mutually
exclusive stream blanking. For any given training sample, at most $N-1$ modalities
are zeroed according to a Bernoulli probability $p(t)$ that scales smoothly from
0 to $\pmax$ over $\Tramp$ epochs:
\begin{equation}
p(t) = \pmax \cdot \min\!\big(1,\; (t + 1) / \Tramp\big)
\end{equation}

With $\pmax=0.48$ and $\Tramp=30$ (our settings), $p(0)\approx2\%$, $p(29)=48\%$,
$p(30+)=48\%$, so blanking is negligible early on and reaches its ceiling only
after integration weights have stabilized. Blanking zeroes the stream input
before the convolution, and gradient flow through that stream is fully detached:
it contributes neither signal nor gradient to the integration pathway or sibling
streams. Three implementation properties guarantee this does not corrupt training
state: (1) gradient detachment: a blanked stream contributes no signal or
gradient to any other stream; (2) BN isolation: batch normalization statistics
are computed exclusively over active samples via index selection, leaving the
blanked zeros unmodified and running statistics uncontaminated; (3) clean
integration arithmetic: $V_i{*}0=0$, so the blanked stream drops algebraically
from $\zint$. We deliberately use zero-masking rather than trainable
missing-modality placeholder tokens (cf.\ \cite{nezakati2024}): we want absence
to be algebraically invisible so that the integration stream learns from whatever
signals are present, rather than developing a learned response to absence
patterns. The resulting stream independence provides inference-time robustness as
a natural consequence. We considered two alternative schedules: static dropout
from epoch 0 and delayed activation after a warm-up phase; high initial dropout
prevents early integration weight formation, while delayed introduction lacks the
plasticity needed to overcome solidified co-dependence. Our progressive schedule
succeeds by ensuring continuous integration forms before independent robustness is
demanded.

We use a mathematically scheduled ramp rather than the adaptive entropy-gated
approaches discussed in Section~\ref{sec:related} because continuous per-layer
fusion introduces a structural incompatibility: in discrete-fusion architectures,
adaptive gating operates at a single well-defined fusion point where auxiliary
classifiers can evaluate semantic confidence. In LINet, fusion occurs at every
layer, and gradients flow from a single classifier through the integration stream
back to the independent streams. Introducing auxiliary classifiers at
intermediate layers would distort this clean gradient path by injecting competing
loss signals into the stream backbones. The only architecturally consistent
alternative, gating based on Layer 1 feature map entropy, measures low-level
spatial statistics (edge density, texture complexity) rather than semantic
confidence, creating a Texture Trap in which a textureless depth map appears
``confident'' despite carrying limited discriminative information. Because a Layer
1 gating decision permanently drops a stream from all subsequent integration
points, the cost of such a misclassification compounds across the full network
depth.

\subsection{Ablative Stream Monitoring}
\label{sec:ablmon}
To monitor pathway utilization without the confounding noise and parameter
overhead of auxiliary classifiers, we employ an \textbf{ablative monitoring}
technique. At the conclusion of each epoch the network is evaluated on a
representative random subset of the samples with stream $S_i$ uniformly zeroed. By
comparing this blanked performance against the standard accuracy achieved during
the unmasked epoch, we define \textbf{Stream Contribution $C_i$} as:
\begin{equation}
C_i = \mathrm{Acc}_{\mathrm{baseline}} - \mathrm{Acc}_{\mathrm{blanked}\_i}
\end{equation}

A disproportionately high $C_i$ for a single stream indicates \textbf{modality
dominance} (lazy routing), whereas uniformly high $C_i$ across all streams reveals
\textbf{pathway collapse}, a state of brittle co-dependence where the integration
pathway fails entirely if any single modality is missing. Conversely, uniformly
low $C_i$ alongside high baseline accuracy indicates robust, redundant
representations. Because this evaluation occurs with gradient updates paused
(inference mode), batch normalization statistics and network weights are
inherently frozen. Furthermore, our clean integration arithmetic ($V_i{*}0=0$)
guarantees that the blanked stream drops cleanly from the pre-activation
summation without requiring architectural branching. This monitoring protocol
directly measures negative co-learning (NCL): uniformly high $C_i$ indicates that
joint training has suppressed unimodal capacity below what independent training
would yield, the defining signature of NCL \cite{magal2025}.

\section{Experiments}

\subsection{Dataset and Evaluation Protocol}
\label{sec:dataset}
We evaluate on \textbf{SUN RGB-D} \cite{songS2015}, which contains 10,335 aligned
RGB-D image pairs captured across multiple sensors. Following standard protocol
\cite{ayub2020, du2019}, we use the 19 scene categories with more than 80 images
each, with significant class imbalance (bedroom 12.6\%, computer room 1.0\%) and
the official split of 4,845 training / 4,659 test images. Images are resized to $256\times256$ and randomly cropped to
$224\times224$, consistent with the preprocessing used by prior work on SUN RGB-D
\cite{ayub2020, du2019}. Depth is provided as raw single-channel
metric values in millimeters, with 0 marking missing pixels. Spatial
transformations, such as random crops and horizontal flips, are applied
synchronously to preserve alignment, while photometric transformations like color
jitter are applied exclusively to the RGB stream. For a complete list of
hyperparameter search spaces and modality-specific augmentation configurations,
please refer to Appendix~A. As a secondary benchmark we additionally evaluate on
\textbf{NYU Depth V2} \cite{silberman2012}, a smaller indoor RGB-D dataset
(1,449 images, 10 scene categories) commonly reported alongside SUN RGB-D, to
confirm that LINet's training dynamics transfer beyond a single dataset rather
than reflecting overfitting to a specific benchmark. We evaluate NYU Depth V2
strictly from scratch, without transferring from a SUN RGB-D-pretrained model, so
that the two benchmarks remain independent tests of the architecture rather than
sharing data across them (SUN RGB-D incorporates NYU Depth V2 imagery, so
evaluating NYU from scratch keeps the two disjoint).

For pretraining experiments we additionally use a 100K-frame subset of
\textbf{ScanNet} \cite{dai2017}, sampled across all available scenes. ScanNet
was selected over synthetic alternatives because its RGB-D pairs are captured with
real consumer-grade depth sensors, preserving the same noise modes (sensor
occlusions, missing values at reflective surfaces, and depth shadows at object
boundaries) present in SUN RGB-D. Although the corpus is small relative to
ImageNet-scale pretraining and its scene labels do not align with SUN RGB-D's
19-class taxonomy, our objective here is not to match Places365-pretrained
baselines absolutely, but to test whether modest in-domain RGB-D pretraining
yields stronger features for downstream fine-tuning. Both streams transfer from
this single pretraining run with no modality-specific initialization, so the
from-scratch and pretrained regimes share one stance in which stream
initialization is never asymmetric across modalities.

\subsection{Implementation Details}
\label{sec:impl}
The LIConv2d operator
restructures the convolutional node by maintaining distinct, parallel weight
matrices per layer, rendering naive transfer of RGB-only ImageNet weights
impossible. Furthermore, as demonstrated by Song, X. et al.\ \cite{songX2017}, fine-tuning
ImageNet-pretrained RGB weights onto depth channels is inherently flawed due to
diverging low-level visual regularities, and Du et al.\ \cite{du2019} showed that
training depth-specific networks from scratch outperforms fine-tuned RGB
transfers. To evaluate our architecture while maintaining fair comparability with
existing literature, we conduct training under two distinct regimes:

\textbf{1. From Scratch.} LINet and all internal baselines are trained strictly on
the SUN RGB-D dataset. This isolates the intrinsic gradient dynamics and routing
capabilities of continuous integration without pre-training biases.

\textbf{2. RGB-D Pre-trained.} To verify that LINet supports similar
transfer-learning workflows to state-of-the-art methods, we additionally pre-train
LINet on a larger RGB-D corpus (ScanNet) before fine-tuning on SUN RGB-D.

LINet's architecture expands horizontally via parallel modality streams rather
than stacking additional sequential convolutions. Utilizing standard ResNet18
backbones ($\sim$11.7M parameters), this three-stream architecture would natively
push the effective parameter count to $\sim$23M. However, against only 4,845
training images, a full-width model of this scale quickly memorizes the training
data, resulting in overfitting. To bring the capacity-to-data ratio into a
workable range, a network width factor $\alpha=0.75$ is applied uniformly to all
stream channels across LINet and our baselines, cutting parameters by $\sim$44\%.
Under this uniform scaling constraint, the single-stream Early Fusion baseline
requires $\sim$6.6M parameters (1.08 GFLOPs). The Late Fusion baseline, which
necessitates duplicate parallel backbones, results in $\sim$12.6M parameters
(2.04 GFLOPs). Finally, the LINet architecture requires $\sim$15.3M parameters
(2.50 GFLOPs).

Hyperparameters were determined through Bayesian optimization (Ray Tune, TPE) with
500 trials using a stratified 20\% validation holdout. Models train using AMP, and
HPO was distributed across NVIDIA A100 GPUs, while the final pretraining and
fine-tuning evaluations were executed on a single NVIDIA T4. We optimize the
cross-entropy classification loss using AdamW (learning rate $1.53\times10^{-4}$,
weight decay $6.74\times10^{-5}$) with a cosine annealing schedule (minimum
threshold $\eta_{\min} = 3.27\times10^{-6}$). To stabilize early spatial
extraction, gradients are clipped at a maximum norm of 1.00. Regularization
includes label smoothing ($\alpha = 0.10$) and a standard dropout ($p = 0.28$).
Importantly, this standard dropout is distinct from Modality Dropout (MD), which
blanks entire modality input streams. Because MD fundamentally alters
the optimization landscape, independent searches were conducted for the baseline
(No-MD) and Progressive MD configurations to ensure a fair comparison. Per-run
optimal values for every reported model are listed in Appendix~B.

\subsection{Results}
\label{sec:results}
Table~\ref{tab:sunrgbd} reports the Mean Class Accuracy (MCA) (consistent with
standard benchmark practice \cite{songS2015, ayub2020}) on the SUN
RGB-D 19-class test split. We evaluate models both trained from scratch and
fine-tuned from large-scale pretraining to isolate the network's architectural
mechanics from external data priors.

\begin{table}[H]
\centering
\footnotesize
\setlength{\tabcolsep}{6pt}
\resizebox{\columnwidth}{!}{%
\begin{tabular}{@{}lcccc@{}}
\toprule
\textbf{Method} & \textbf{Pretrained} & \textbf{RGB} & \textbf{Depth} & \textbf{Fusion} \\
\midrule
ResNet18 baseline               & No        & 36.0          & 37.4          & ---           \\
ResNet18 early fusion baseline  & No        & ---           & ---           & 39.7          \\
ResNet18 late fusion baseline   & No        & ---           & ---           & 41.7          \\
SS-CNN \cite{liao2015}      & No        & \textbf{36.1} & ---           & 41.3          \\
CoMAE \cite{yang2023}$\dagger$ & No     & 27.5          & \textbf{38.6} & 40.5          \\
\textbf{LINet (no MD)}        & No        & 15.0          & 14.4          & 43.5          \\
\textbf{LINet + prog.\ MD}    & No        & 33.7          & 35.7          & \textbf{45.2} \\
\midrule
Wang, A. et al.\ \cite{wangA2016}         & Places    & 40.4          & 36.5          & 48.1          \\
CoMAE \cite{yang2023}$\S$               & SUN RGB-D & ---           & ---           & 55.2          \\
Du et al.\ \cite{du2019} Aug.          & ImageNet  & \textbf{50.6} & \textbf{47.9} & 56.7          \\
CBCL \cite{ayub2020}     & Places365 & 48.8          & 37.3          & \textbf{59.5} \\
\textbf{LINet (no MD)}        & ScanNet$\ddagger$ & 36.5  & 37.2          & 48.0          \\
\textbf{LINet + prog.\ MD}    & ScanNet$\ddagger$ & 39.2  & 38.4          & 49.6          \\
\bottomrule
\end{tabular}%
}
\caption{Scene classification mean class accuracy (\%) on SUN RGB-D 19-class,
official test split. $\dagger$ CoMAE \cite{yang2023} uses a ViT-B backbone roughly $5.6\times$ the size of LINet ($\sim$86M vs.\ $\sim$15.3M). $\ddagger$ ScanNet pretraining uses only a $\sim$100K-frame subset, far below the Places365/ImageNet corpora; included to show transferability, not as full-scale pretraining. $\S$ Self-supervised on the SUN RGB-D training set, no external corpus.}
\label{tab:sunrgbd}
\end{table}

When trained from scratch, the unregularized LINet architecture (No-MD) achieves a
fusion accuracy of 43.5\%, outperforming both the Early (39.7\%) and Late (41.7\%)
fusion baselines. However, evaluating the isolated pathways reveals severe
modality co-dependence, with the independent RGB and Depth streams degrading to
15.0\% and 14.4\% respectively. The network learns to rely entirely on the joint
distribution at the integration node.

We originally introduced Modality Dropout to force independent stream
representations, operating under the assumption that restricting cross-modal
visibility would inherently penalize the overall fusion score. Surprisingly, the
progressive dropout curriculum not only rescued the isolated stream accuracies
(improving to 33.7\% RGB and 35.7\% Depth), but simultaneously pushed the overall
fusion MCA to 45.2\%, a 1.7pp improvement over the unregularized baseline. This is
the strongest from-scratch result in Table~\ref{tab:sunrgbd}, exceeding even CoMAE
\cite{yang2023}, a self-supervised ViT-B trained from scratch on SUN RGB-D
(40.5\% MCA), despite LINet using roughly a fifth of the parameters ($\sim$15.3M
vs.\ $\sim$86M). Forcing each backbone to function as a competent, independent
feature extractor provides the integration stream with higher-quality inputs. The
learned V weights can leverage stronger signals, enabling more effective
cross-modal routing at every integration layer. We investigate this effect and the
role of schedule design in detail in Section~\ref{sec:ablation}.

ScanNet pretraining transfers: fine-tuned LINet improves from 45.2\%
(from-scratch) to 49.6\% MCA, a $+4.4$pp gain demonstrating that even modest
in-domain RGB-D pretraining produces stronger features for SUN RGB-D. The
remaining gap to Places365-pretrained methods (56.7--59.5\%) reflects three
factors: ScanNet is captured with a single sensor (Structure Sensor) versus SUN
RGB-D's four (Kinect v1/v2, RealSense, Xtion), the 100K-frame subset is small and
label-misaligned relative to the multi-million-image scene-classification corpora
those methods use, and SUN RGB-D's official test split exhibits class-structured
distribution shift affecting all methods on this benchmark
(Section~\ref{sec:conclusion}). CoMAE's pretrained 55.2\% is the most directly
comparable result, as it is the only other entry that does not engineer a
depth-specific stream initialization, instead treating both modalities uniformly
through a shared self-supervised encoder. The Places365 gap therefore reflects the
pretraining corpus, not the fusion mechanism, leaving the from-scratch comparison
the cleaner measure of the architecture itself.

\begin{table}[H]
\centering
\footnotesize
\setlength{\tabcolsep}{6pt}
\resizebox{\columnwidth}{!}{%
\begin{tabular}{@{}lcccc@{}}
\toprule
\textbf{Method} & \textbf{Pretrained} & \textbf{RGB} & \textbf{Depth} & \textbf{Fusion} \\
\midrule
ResNet18 baseline               & No & \textbf{40.4} & 40.2          & ---           \\
ResNet18 early fusion baseline  & No & ---           & ---           & 44.8          \\
ResNet18 late fusion baseline   & No & ---           & ---           & 49.3          \\
CoMAE \cite{yang2023}$\dagger$ & No & ---        & ---           & 40.7          \\
\textbf{LINet (no MD)}        & No & 27.0          & 16.1          & 50.1          \\
\textbf{LINet + prog.\ MD}    & No & 37.8          & \textbf{40.3} & \textbf{50.8} \\
\bottomrule
\end{tabular}%
}
\caption{Scene classification mean class accuracy (\%) on NYU Depth V2.
$\dagger$ CoMAE \cite{yang2023} is a from-scratch ViT-B model ($\sim$86M
parameters, $\sim$5.6$\times$ LINet); its NYU paper reports only the fused
result.}
\label{tab:nyu}
\end{table}

NYU Depth V2 reproduces the from-scratch SUN RGB-D pattern. The unregularized
LINet (No-MD) achieves 50.1\% fusion MCA but exhibits the same severe modality
co-dependence: RGB-only collapses to 27.0\% and Depth-only to 16.1\%. Progressive
modality dropout reverses the co-dependence (RGB 37.8\%, Depth 40.3\%) and
simultaneously improves fusion MCA to 50.8\%, outperforming both the early-fusion
(44.8\%) and late-fusion (49.3\%) ResNet18 baselines.

\subsection{Ablation Studies and Optimization Dynamics}
\label{sec:ablation}
To examine how Modality Dropout schedules interact with the continuous integration
mechanism of LINet, we conduct targeted ablation studies across four training
conditions and two dropout intensities. We first validate the co-dependence
reversal claimed in Section~\ref{sec:results} and compare schedule performance
(Section~\ref{sec:streamhealth}), then examine the internal routing dynamics that
explain the performance differences (Section~\ref{sec:weightdyn}), and finally
analyze how per-channel modality routing emerges from the interaction between
symmetric integration weights and asymmetric input distributions
(Section~\ref{sec:routing}).

\subsubsection{Stream Health, Schedule Comparison, and Architectural Robustness}
\label{sec:streamhealth}
\begin{figure}[H]
\centering
\includegraphics[width=\columnwidth]{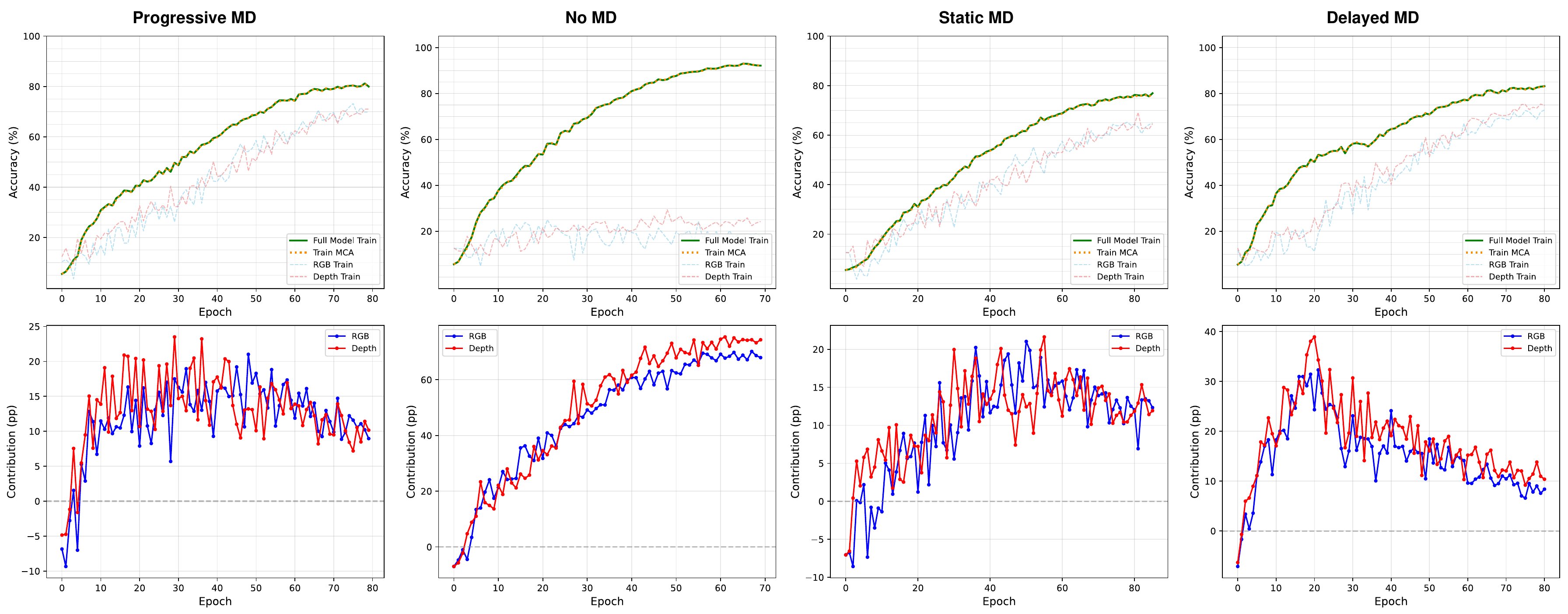}
\caption{Training accuracy and stream contribution ($C_i$) across four dropout
schedules. Without dropout, the full model reaches $\sim$95\% training
accuracy but individual streams stagnate at $\sim$20\%, producing contributions of
75--80pp, the signature of severe pathway collapse. All three MD variants produce
balanced, low contributions (9--15pp), reversing the co-dependence.}
\label{fig:training_curves}
\end{figure}

Section~\ref{sec:results} reported that the unregularized architecture suffers
from severe modality co-dependence while progressive dropout restores independent
stream capacity. Figure~\ref{fig:training_curves} validates this claim, and
reveals that the pattern holds across all four dropout schedules, by visualizing
the training dynamics under each condition using the ablative monitoring protocol
defined in Section~\ref{sec:ablmon}.

The contrast between the unregularized baseline and any form of dropout is stark.
Without dropout (Figure~\ref{fig:training_curves}), the network achieves
near-perfect training accuracy (95\%) but the individual streams are functionally
useless in isolation: stream contribution scores of $\CRGB \approx 75$pp and
$\CDepth \approx 80$pp indicate that removing either modality causes an accuracy
collapse of 75--80 percentage points. This is the defining signature of negative
co-learning \cite{magal2025}, in which joint training actively suppresses
unimodal capacity below what independent training would yield. Under any MD
schedule (Figure~\ref{fig:training_curves}), this co-dependence reverses:
individual streams reach training accuracies of 55--65\%, and $C_i$ scores drop to
9--15pp, indicating robust, redundant representations.

\begin{table}[H]
\centering
\footnotesize
\setlength{\tabcolsep}{8pt}
\resizebox{\columnwidth}{!}{%
\begin{tabular}{@{}lcccccc@{}}
\toprule
\textbf{Schedule} & \boldmath$\pmax$ & \textbf{Fusion} & \textbf{RGB} & \textbf{Depth} & \boldmath$\CRGB$ & \boldmath$\CDepth$ \\
\midrule
No MD                       & ---  & 43.5          & 15.0 & 14.4 & $\sim$75 & $\sim$80 \\
Static MD                   & 0.48 & 43.2          & 32.9 & 33.7 & $\sim$11 & $\sim$11 \\
Delayed MD                  & 0.48 & 43.8          & 33.7 & 36.5 & $\sim$9  & $\sim$12 \\
\textbf{Progressive MD}     & 0.48 & \textbf{45.2} & 33.7 & 35.7 & $\sim$12 & $\sim$13 \\
\textit{Progressive MD}$\dagger$ & 0.8 & \textit{44.1} & \textit{35.1} & \textit{37.3} & $\sim$8 & $\sim$9 \\
\bottomrule
\end{tabular}%
}
\caption{Modality Dropout schedule comparison on SUN RGB-D 19-class (MCA \%).
$\dagger$Uses the Progressive schedule with elevated $\pmax = 0.8$ (see text).}
\label{tab:schedules}
\end{table}

All three MD variants rescue stream independence: RGB-only MCA improves from
15.0\% to 32--34\%, Depth-only from 14.4\% to 34--37\%. Crucially, this stream
independence does not come at the cost of fusion performance: every MD schedule
also improves the fusion MCA over the unregularized baseline (43.5\%). The
progressive schedule achieves the highest fusion accuracy at 45.2\%, followed by
Delayed (43.8\%) and Static (43.2\%).

The pattern across all four schedules reveals a consistent finding that challenges
established assumptions about when and how modality dropout can be applied to
multi-stream networks trained from scratch.

\textbf{Static dropout from epoch 0 is not catastrophic.} The multimodal
literature has identified static modality dropout applied from epoch 0 as
catastrophic for from-scratch training, citing gradient starvation that prevents
early feature learning \cite{neverova2015}. Yet LINet achieves 43.2\% fusion
MCA under static dropout, only 2pp below the progressive optimum, with balanced
stream contributions ($\sim$11pp each) comparable to the progressive schedule. The
streams learn robust, independent features despite never receiving a dropout-free
warm-up phase.

\textbf{Delaying dropout does not yield stronger features and risks destabilizing
gradients.} The Delayed schedule ($p=0$ for 20 epochs, then a 15-epoch ramp to
$\pmax$) achieves 43.8\% fusion MCA with the strongest Depth-only accuracy of the
standard configurations (36.5\%). However, the warm-up phase does not translate
into stronger fused representations. As the routing dynamics in
Section~\ref{sec:weightdyn} will show (Figure~\ref{fig:stream_weights}), the
mid-training activation of dropout delivers a regularization shock that disrupts
both backbone and integration weight
trajectories, introducing perturbations that the network cannot fully recover from
before training concludes.

\textbf{Even aggressive dropout ($\pmax=0.8$) remains viable from scratch.}
Motivated by Magal et al.\ \cite{magal2025}, who demonstrated that aggressive static dropout
rates (up to $p=0.8$) fundamentally alter co-learning dynamics and required heavy
pretraining to remain viable, we evaluated the progressive schedule with
$\pmax=0.8$. LINet achieves 44.1\% fusion MCA (trained entirely from scratch,
without any pretraining), producing the strongest unimodal performance of any
configuration (RGB 35.1\%, Depth 37.3\%).

These results point to a property of LINet that discrete-fusion architectures do
not share: \textbf{inherent architectural robustness to modality dropout.} We
attribute this to two properties of continuous integration. First, the $1/N$
constant initialization of the bridging weights (Section~\ref{sec:init}) provides
coherent, stable gradient flow from the first update, preventing the gradient
starvation that makes static dropout catastrophic when bridging weights are
Kaiming-initialized. Second, the integration stream acts as a decision-maker that
combines modality signals via the learned V weights. Without MD, the streams
become co-dependent: each learns only to provide its ``half'' of a joint pattern.
With MD, each stream must independently carry a discriminative signal. The
integration stream then receives two independently strong inputs, which gives V
weights richer, more redundant signals to route, yielding higher fusion accuracy,
and even when one stream is dropped, the surviving modality still feeds every
integration point in the network. The slight decrease in fusion MCA from
$\pmax=0.48$ (45.2\%) to $\pmax=0.8$ (44.1\%), even as unimodal accuracies increase
monotonically, reveals a Pareto frontier: at extreme dropout rates, the
integration stream sees paired inputs too infrequently to fully optimize
cross-modal routing, even as the individual streams strengthen. The full
characterization of this trade-off we leave to future work.

\subsubsection{Weight Dynamics Across Schedules}
\label{sec:weightdyn}
Section~\ref{sec:streamhealth} established that all three MD schedules produce
comparably robust independent streams, yet fusion accuracy differs: Progressive
(45.2\%) outperforms Delayed (43.8\%) and Static (43.2\%). To examine how these
schedules affect the network internally, we track weight dynamics at two levels:
the stream backbone weights ($W_1, W_2$), which reflect how each modality's
feature extractor develops, and the integration bridging weights
($V_1, V_2$), which reflect how the integration
mechanism combines the two modalities at each layer.

\begin{figure}[H]
\centering
\includegraphics[width=\columnwidth]{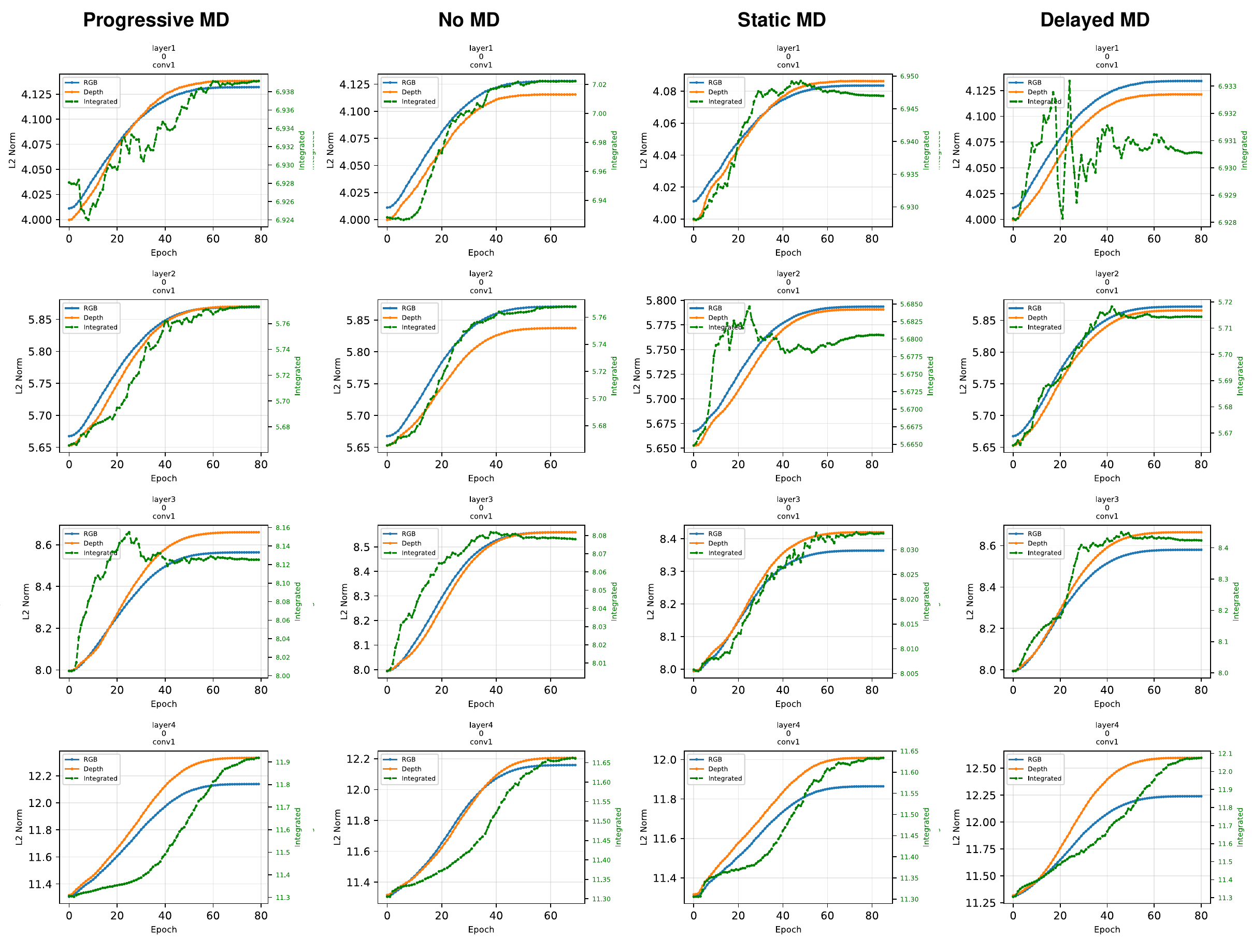}
\caption{Stream backbone weight L2 norms (blue=RGB, orange=Depth,
green=Integrated) across four network layers (conv1). No-MD: RGB and Depth
weights diverge in deeper layers, with RGB growing faster. Under all
three MD schedules, backbone weights evolve in close parallel. Note the visible
perturbation in Delayed MD layer 1 around epoch 30, when dropout activates.}
\label{fig:stream_weights}
\end{figure}

Figure~\ref{fig:stream_weights} reveals a consistent pattern: under all three MD
schedules, the RGB and Depth backbone weights grow in close parallel across
all network depths. The feature extractors develop balanced capacity regardless of
whether dropout is applied from epoch 0, after a delay, or with a progressive
ramp. Only the unregularized baseline exhibits backbone divergence: in layers
3--4, RGB weights grow measurably faster than Depth, confirming that without
dropout the network allocates more representational capacity to RGB, consistent
with the modality dominance reported in Table~\ref{tab:schedules}.

The Delayed schedule (Figure~\ref{fig:stream_weights}) shows a distinctive
perturbation in layer 1 around epoch 30, when dropout activates. The Depth backbone
weights visibly destabilize as the optimizer adjusts to the sudden introduction of
missing-modality inputs. This is a concrete manifestation of the regularization
shock identified in Section~\ref{sec:streamhealth}: the mid-training activation of
dropout disrupts feature learning dynamics that had already settled into a stable
trajectory.

Backbone balance is the load-bearing factor. As Section~\ref{sec:routing} will
show, the integration mechanism's per-channel modality routing emerges from the
interaction between symmetric integration weights and stream activations; balanced
backbones produce balanced activations, which the routing mechanism preserves into
the integrated stream. What distinguishes Progressive is not a unique routing
pattern but the absence of disruption: it is the only schedule that never
experiences a phase transition (no shock from delayed activation, no abrupt
exposure to dropout), allowing the co-adaptation between $W_1$, $W_2$, and the
integration weights to proceed smoothly across the full training trajectory. This
is the temporal analogue of the initialization principle from
Section~\ref{sec:init}: just as the V weights must start with coherent gradients
($1/N$ init) rather than random noise, the dropout schedule must start with
coherent cross-modal signal (low $p$) rather than heavy masking. Progressive
achieves both by ensuring the optimizer never encounters a discontinuity in the
training objective.

\subsubsection{Per-Channel Modality Routing in Continuous Integration}
\label{sec:routing}
We find that LINet's continuous integration mechanism produces genuine per-channel
modality preference at the activation level, with 5--26\% of integration output
channels showing $>2\times$ preference for one modality across layers (peaking at
conv1 with 35\% of channels). Notably, this asymmetry is not expressed through
asymmetric weights. Per-channel norm ratios
$\|V_1[k,:]\| / \|V_2[k,:]\|$ across all 20 LIConv2d layers cluster tightly around
1 (median 0.97--1.06, IQR of $\log_{10}(\mathrm{ratio}) < 0.13$, no channels with
$>2\times$ dominance), $V_1$ and $V_2$ are matrix-wise uncorrelated (vectorized
cosine alignment $\approx 0$), and stream backbones $W_1$ and $W_2$ show the same
row-norm symmetry while learning distinct projections per modality. The
per-channel modality preference instead emerges from the interaction between
symmetric weights and asymmetric input distributions: each modality's distinctive
feature statistics (color/texture variance for RGB, geometric variance for Depth)
propagate through their respective backbones, producing per-channel activations
that vary in modality dominance even when the V weights themselves are matched in
magnitude. The integrated stream then mixes these data-driven modality asymmetries
through its self-recurrence ($\Vprev$), batch normalization, and downstream
LIConv2d layers. The architecture performs per-channel modality routing,
expressing it through the interaction between symmetric weights and asymmetric
input distributions rather than through asymmetric weights.

\section{Conclusion}
\label{sec:conclusion}
LINet demonstrates that the historical early-vs-late fusion dilemma can be
effectively addressed through continuous, per-layer pre-activation integration,
bypassing the need for empirically placed intermediate fusion blocks. The LIConv2d
operator maintains dedicated per-modality spatial filters and combines their
pre-activation outputs, achieving modality specialization and cross-modal learning
simultaneously. Our findings highlight two key considerations for continuous
multi-stream networks. First, integration mechanism initialization is a critical
structural decision: because bridging $1\times1$ convolutions occupy a privileged
backward-pass position, applying a $1/N$ constant initialization proved necessary
in our architecture to avoid early gradient corruption. Second, progressive
modality dropout provides an effective curriculum for preventing pathway collapse,
a form of negative co-learning in which joint training suppresses unimodal
capacity. By allowing integration to form before demanding independent pathway
robustness, our results indicate that the temporal shape of the dropout schedule
can be a critical design factor alongside the dropout rate.

\noindent\textbf{Limitations:} We observe substantial within-dataset distribution
shift in SUN RGB-D's official test split that affects benchmark interpretability
for all methods evaluated on this dataset, not LINet specifically.

Per-class analysis reveals distribution shifts within SUN RGB-D's official split
that are invisible at the aggregate level. While overall sensor proportions match
across train/val/test ($\approx$19/37/11/32\% for Kinect v1/v2, RealSense, Xtion),
per-class sensor distributions shift substantially: classroom is 73\% Kinect v2 on
test vs.\ 31\% in training; office is 40\% Xtion vs.\ 14\%; dining\_area is 56\%
Kinect v2 vs.\ 39\%. For library, where aggregate sensor proportions are stable,
the failure is driven by a subcollection shift within RealSense (sa/* recordings
in training vs.\ lg/* in test). Per-sensor test MCA spans an 11.7pp range across
the four sensors, and eight rep-drift classes (dining\_area, discussion\_area,
library, classroom, lab, lecture\_theatre, home\_office, study\_space) show
5--30\% nearest-neighbor match between test and training features in
penultimate-layer space, compared to 73\% on stable classes, with centroid drift
of 0.4--0.7. Adaptive batch normalization at inference recovers $+4.95$pp MCA on
these rep-drift classes, confirming that batch-statistic mismatch contributes to
the gap. These patterns suggest SUN RGB-D's official split stratifies by recording
campaign rather than uniformly, producing class-specific distribution shifts
undetectable without per-class breakdowns.

The dominant per-class confusions on test overlap with validation confusions for
three of the five failing classes (library, office, classroom: 2/3 top-3 overlap),
with partial overlap for bedroom (1/3) and none for dining\_area (0/3), indicating
that the test distribution for dining\_area contains visual subtypes not
represented in validation: top test scenes (e.g.\ mit\_w85\_basement, mit\_lab\_hj,
mit\_46\_4conf\_2) suggest dining environments embedded in academic buildings,
distinct from the bar and cafeteria contexts present in training. Aggregate
feature-distribution shift is modest (whole-split RBF MMD$^2 = 0.010$
train$\leftrightarrow$test) and per-class feature drift correlates weakly with
per-class accuracy gap (Spearman $\rho = 0.37$, $p = 0.13$), consistent with
class-structured shifts that aggregate statistics miss. Prior work \cite{ayub2020}
reports improved SUN RGB-D test performance using methods such as
centroid-based classification in place of softmax. The within-class heterogeneity
our diagnostics expose suggests the classifier formulation is one factor in the
gap, though identifying the operative mechanism remains a direct ablation left for
future work.

\onecolumn
\renewcommand{\thetable}{B\arabic{table}}
\setcounter{table}{0}

\section*{Appendix A: Training Configuration}
This appendix provides full training configurations for the from-scratch and
pretrained runs reported in Section~\ref{sec:results}. All values are the optimum
returned by the HPO campaign described below; for hyperparameters reported as a
single point estimate, separate independent searches were conducted for the
No-MD baseline and Progressive MD configuration to ensure fair comparison.

\textbf{Hyperparameter Search.} Bayesian optimization with the Tree-structured
Parzen Estimator (TPE) implementation in Ray Tune, 500 trials per training regime
against a stratified scene-grouped 20\% validation holdout. The search spanned
learning rate and weight decay (log-uniform), label smoothing, standard dropout,
gradient clip max-norm, and the per-modality augmentation dials; the optimum
returned for each reported run is given in the tables below. Modality Dropout
configurations searched $\pmax$ and $\Tramp$ independently from the No-MD baseline
since the dropout schedule fundamentally alters the optimization landscape.

\textbf{Optimization.} Representative configuration (from-scratch Progressive MD;
per-run optima in Appendix B): AdamW (lr $= 1.53\times10^{-4}$, wd
$= 6.74\times10^{-5}$), cosine annealing to $\eta_{\min} = 3.27\times10^{-6}$,
gradient clip max-norm 1.00, label smoothing $\alpha = 0.10$, standard dropout
$p = 0.28$, stem-LR multiplier 21, cross-entropy classification loss,
mixed-precision (AMP) training. Modality Dropout: progressive linear ramp
$p(t) = 0.48 \cdot \min(1, (t+1)/30)$ with $\Tramp = 30$ epochs. The single values
quoted above are representative; the exact per-configuration optima for every
reported model are tabulated in Appendix B (Tables~\ref{tab:b1}--\ref{tab:b5}).

\textbf{Training Schedule.} From-scratch (SUN RGB-D): 70--86 epochs (No-MD 70,
Progressive MD 80; see Appendix B), batch size 64, single NVIDIA T4. Pretrained
fine-tuning (SUN RGB-D): best checkpoint at epoch 13 (No-MD) / 18 (Progressive
MD), batch size 64, single NVIDIA T4. ScanNet pretraining: best checkpoint at
epoch 50 over 100K frames, batch size 128, NVIDIA A100. HPO trials: distributed
across NVIDIA A100 GPUs.

\textbf{Augmentation.} Spatial transformations (random crop $224\times224$ from
$256\times256$, horizontal flip) applied synchronously across both streams to
preserve modality alignment. RGB-only photometric augmentations: color jitter
(brightness, contrast, saturation, hue), random grayscale, Gaussian blur, random
erasing. Depth-only augmentations: scale jitter, Gaussian blur, hole dropout.
Augmentation probabilities and magnitudes were jointly tuned with all other
hyperparameters in HPO. Rather than searching an independent probability and
magnitude for each of the dozen-plus augmentations, each modality exposes two
global control dials, an aggregate probability scale and an aggregate magnitude
scale, that jointly modulate all of that modality's augmentations. The per-run
``aug prob'' and ``aug mag'' values in Appendix B are these dial settings, not
literal probabilities; values above 1.0 simply turn the corresponding dial past
its nominal setting.

\textbf{Width Scaling.} Network width factor $\alpha = 0.75$ applied uniformly to
all stream channels (LINet and baselines). Resulting channel progression at the
four ResNet18 stages: $[48, 48, 96, 192, 384]$.

\section*{Appendix B: Per-Configuration Hyperparameters}
This appendix lists the exact optimal hyperparameters returned by the HPO
campaign (Section~\ref{sec:impl}, Appendix A) for every model reported in
Tables~\ref{tab:sunrgbd}--\ref{tab:schedules}. Each configuration was searched
independently; values below are the per-run optima rather than a shared default.
All from-scratch and fine-tuning runs use AdamW with cosine annealing,
cross-entropy loss, and mixed-precision training at width factor $\alpha = 0.75$.
The ``RGB/Depth aug prob'' and ``aug mag'' rows are the two per-modality control
dials defined in Appendix A, not literal per-augmentation probabilities; values
above 1.0 indicate a dial turned past its nominal setting. Batch size is 64 for
all SUN RGB-D and NYU runs and 128 for ScanNet pretraining.

\begin{table}[H]
\centering
\footnotesize
\begin{threeparttable}
\caption{SUN RGB-D (from scratch): LINet across four modality-dropout schedules.}
\label{tab:b1}
\begin{tabular}{@{}lcccc@{}}
\toprule
\textbf{Hyperparameter} & \textbf{No-MD} & \textbf{Progressive MD} & \textbf{Static MD} & \textbf{Delayed MD} \\
\midrule
Epochs $\dagger$            & 70   & 80   & 86   & 81   \\
Learning rate               & 1.57e-4 & 1.53e-4 & 1.15e-4 & 1.53e-4 \\
Weight decay                & 3.52e-4 & 6.74e-5 & 9.22e-5 & 1.46e-4 \\
$\eta_{\min}$ (cosine)      & 3.94e-6 & 3.27e-6 & 3.48e-6 & 1.73e-6 \\
Dropout $p$                 & 0.38 & 0.28 & 0.34 & 0.27 \\
Label smoothing $\alpha$    & 0.18 & 0.10 & 0.10 & 0.10 \\
Grad-clip max-norm          & 1.00 & 1.00 & 1.00 & 1.00 \\
Stem-LR multiplier          & 13   & 21   & 21   & 21   \\
RGB aug prob                & 0.73 & 0.94 & 0.97 & 0.92 \\
RGB aug mag                 & 1.01 & 1.20 & 1.10 & 1.17 \\
Depth aug prob              & 0.79 & 1.06 & 1.06 & 0.95 \\
Depth aug mag               & 1.08 & 0.98 & 0.94 & 1.04 \\
MD $\pmax$                  & ---  & 0.48 & 0.48 & 0.48 \\
MD schedule                 & none & ramp 0--30 ep & constant from ep 0 & start ep 20, ramp 15 \\
Test MCA (F / RGB / D)      & 43.5 / 15.0 / 14.4 & 45.2 / 33.7 / 35.7 & 43.2 / 32.9 / 33.7 & 43.8 / 33.7 / 36.5 \\
Test Acc (F / RGB / D)      & 49.2 / 15.0 / 17.4 & 49.4 / 36.4 / 37.8 & 47.2 / 36.1 / 36.0 & 48.8 / 37.0 / 40.1 \\
\bottomrule
\end{tabular}
\begin{tablenotes}[flushleft]\footnotesize
\item[$\dagger$] All runs use a fixed epoch budget equal to the best epoch found on an
80/20 train/val split of the SUN training set (the official SUN split provides
only train and test); the train/test run then retrains for exactly that many
epochs with no validation or best-weight restoration, all other settings
(including $T_{\max}$) held constant. Static MD and Delayed MD share the
Progressive-MD configuration except for the dropout schedule. The Progressive-MD
($\pmax = 0.8$) variant in Table~\ref{tab:schedules} (Fusion 44.1) uses the
Progressive-MD column verbatim with only $\pmax$ changed to 0.8.
\end{tablenotes}
\end{threeparttable}
\end{table}

\begin{table}[H]
\centering
\footnotesize
\begin{threeparttable}
\caption{SUN RGB-D (from scratch): single-stream and fusion baselines.}
\label{tab:b2}
\begin{tabular}{@{}lcccc@{}}
\toprule
\textbf{Hyperparameter} & \textbf{RGB-only} & \textbf{Depth-only} & \textbf{Early fusion} & \textbf{Late fusion} \\
\midrule
Best epoch                  & 65 & 59 & 65 & 43 \\
Learning rate               & 4.91e-4 & 5.07e-4 & 7.51e-4 & 2.67e-4 \\
Weight decay                & 2.36e-3 & 1.64e-3 & 1.78e-3 & 7.81e-4 \\
$\eta_{\min}$ (cosine)      & 6.91e-6 & 7.49e-6 & 6.77e-6 & 5.77e-6 \\
Dropout $p$                 & 0.38 & 0.39 & 0.37 & 0.39 \\
Label smoothing $\alpha$    & 0.20 & 0.20 & 0.16 & 0.19 \\
Grad-clip max-norm          & 1.30 & 1.25 & 1.30 & 1.45 \\
RGB aug prob                & 0.82 & 1.00 & 0.86 & 0.87 \\
RGB aug mag                 & 0.92 & 1.00 & 0.94 & 0.98 \\
Depth aug prob              & 1.00 & 0.97 & 0.94 & 0.97 \\
Depth aug mag               & 1.00 & 0.83 & 0.89 & 0.86 \\
Test MCA (Table~\ref{tab:sunrgbd}) & 36.0 & 37.4 & 39.7 & 41.7 \\
\bottomrule
\end{tabular}
\begin{tablenotes}[flushleft]\footnotesize
\item Stem-LR multiplier does not apply to single-stream, early-, or late-fusion
baselines. Test MCA values are reproduced from Table~\ref{tab:sunrgbd}.
\end{tablenotes}
\end{threeparttable}
\end{table}

\begin{table}[H]
\centering
\footnotesize
\begin{threeparttable}
\caption{SUN RGB-D: ScanNet-pretrained LINet (fine-tuning phase).}
\label{tab:b3}
\begin{tabular}{@{}lcc@{}}
\toprule
\textbf{Hyperparameter} & \textbf{No-MD} & \textbf{Progressive MD} \\
\midrule
Best epoch                  & 13 & 18 \\
Best val MCA                & 62.6\% & 63.8\% \\
Learning rate               & 1.884e-4 & 1.404e-4 \\
Weight decay                & 4.138e-2 & 7.293e-4 \\
$\eta_{\min}$ (cosine)      & 4.724e-6 & 9.632e-7 \\
Dropout $p$                 & 0.78 & 0.65 \\
Label smoothing $\alpha$    & 0.05 & 0.05 \\
Grad-clip max-norm          & 1.00 & 1.20 \\
Stem-LR multiplier          & 1.70 & 0.90 \\
RGB aug prob                & 0.85 & 0.70 \\
RGB aug mag                 & 1.20 & 1.07 \\
Depth aug prob              & 1.27 & 1.26 \\
Depth aug mag               & 1.27 & 1.10 \\
MD $\pmax$                  & --- & 0.48 \\
Test MCA (F / RGB / D)      & 48.0 / 36.5 / 37.2 & 49.6 / 39.2 / 38.4 \\
Test Acc (F / RGB / D)      & 49.2 / 38.0 / 38.7 & 53.1 / 40.5 / 42.2 \\
\bottomrule
\end{tabular}
\begin{tablenotes}[flushleft]\footnotesize
\item Values are for the SUN RGB-D fine-tuning phase. The ScanNet pretraining phase
(shared across the No-MD and Progressive-MD pretraining runs; 100K-frame subset,
batch size 128, NVIDIA A100, best checkpoint at epoch 50) used: lr
$3.14\times10^{-4}$, wd $4.01\times10^{-4}$, $\eta_{\min}$ $3\times10^{-6}$,
dropout 0.13, label smoothing 0.07, grad-clip 1.20, stem-LR multiplier 16.0,
rgb\_aug (prob 0.43, mag 0.80), depth\_aug (prob 0.54, mag 0.70); the only
difference between the two pretraining runs is the MD schedule. On ScanNet itself,
pretraining-phase MCA (both / RGB-only / Depth-only) was 99.8 / 94.6 / 27.4 for
No-MD and 99.8 / 98.9 / 97.3 with Progressive MD: both reach near-ceiling fused
accuracy, but only with MD does the depth-only stream recover
(27.4 $\rightarrow$ 97.3), the same co-dependence reversal observed on SUN.
\end{tablenotes}
\end{threeparttable}
\end{table}

\begin{table}[H]
\centering
\footnotesize
\begin{threeparttable}
\caption{NYU Depth V2: LINet (from scratch).}
\label{tab:b4}
\begin{tabular}{@{}lcc@{}}
\toprule
\textbf{Hyperparameter} & \textbf{No-MD} & \textbf{Progressive MD} \\
\midrule
Best epoch                  & 61 & 65 \\
Learning rate               & 1.40e-4 & 1.40e-4 \\
Weight decay                & 9.07e-4 & 1.36e-4 \\
$\eta_{\min}$ (cosine)      & 3.84e-6 & 2.17e-6 \\
Dropout $p$                 & 0.41 & 0.41 \\
Label smoothing $\alpha$    & 0.19 & 0.16 \\
Grad-clip max-norm          & 1.35 & 1.05 \\
Stem-LR multiplier          & 20 & 23 \\
RGB aug prob                & 1.13 & 0.99 \\
RGB aug mag                 & 1.13 & 0.87 \\
Depth aug prob              & 0.71 & 0.75 \\
Depth aug mag               & 0.91 & 0.80 \\
MD $\pmax$                  & --- & 0.44 \\
Test MCA (F / RGB / D)      & 50.1 / 27.0 / 16.1 & 50.8 / 37.8 / 40.3 \\
Test Acc (F / RGB / D)      & 49.5 / 25.8 / 26.8 & 49.5 / 30.9 / 38.8 \\
\bottomrule
\end{tabular}
\end{threeparttable}
\end{table}

\begin{table}[H]
\centering
\footnotesize
\begin{threeparttable}
\caption{NYU Depth V2: single-stream and fusion baselines.}
\label{tab:b5}
\begin{tabular}{@{}lcccc@{}}
\toprule
\textbf{Hyperparameter} & \textbf{RGB-only} & \textbf{Depth-only} & \textbf{Early fusion} & \textbf{Late fusion} \\
\midrule
Epochs $\ddagger$           & 67 & 71 & 36 & 86 \\
Learning rate               & 1.41e-4 & 1.53e-4 & 1.74e-4 & 1.82e-4 \\
Weight decay                & 7.16e-4 & 4.56e-5 & 7.95e-4 & 1.86e-4 \\
$\eta_{\min}$ (cosine)      & 1.71e-6 & 2.67e-6 & 1.07e-6 & 2.66e-6 \\
Dropout $p$                 & 0.36 & 0.43 & 0.45 & 0.41 \\
Label smoothing $\alpha$    & 0.17 & 0.17 & 0.18 & 0.16 \\
Grad-clip max-norm          & 1.00 & 1.15 & 1.00 & 1.10 \\
RGB aug prob                & 1.12 & 1.00 & 1.20 & 1.03 \\
RGB aug mag                 & 0.91 & 1.00 & 0.80 & 0.94 \\
Depth aug prob              & 1.00 & 0.75 & 0.90 & 0.81 \\
Depth aug mag               & 1.00 & 0.83 & 0.92 & 0.83 \\
Val MCA                     & 62.3\% & 59.0\% & 68.4\% & 69.5\% \\
Val Acc                     & 57.9\% & 58.5\% & 59.8\% & 65.4\% \\
\bottomrule
\end{tabular}
\begin{tablenotes}[flushleft]\footnotesize
\item[$\ddagger$] Total training epochs (Early fusion: best checkpoint at epoch 29 of 36).
Baselines report HPO validation MCA / accuracy; the corresponding test MCA
appears in Table~\ref{tab:nyu}.
\end{tablenotes}
\end{threeparttable}
\end{table}

\end{document}